\documentclass[manuscript,screen]{acmart}
\AtBeginDocument{%
  }

\acmJournal{TOMM}

\usepackage{svg}
\usepackage{xcolor}
\usepackage{pifont}
\usepackage{adjustbox}
\usepackage{multirow}
\usepackage[ruled,vlined]{algorithm2e}
\usepackage[table]{xcolor}
\usepackage{booktabs}  
\usepackage{tabularx}  

\begin{document}

\title{SVGS: Single-View to 3D Object Editing via Gaussian Splatting}

\author{Pengcheng Xue}
\affiliation{
  \institution{School of Computer Science and Technology, Zhejiang Gongshang University, Hangzhou}
  \country{China}
}

\author{Yan Tian}
\authornote{Corresponding author: tianyan@zjgsu.edu.cn.}
\affiliation{
  \institution{School of Computer Science and Technology, Zhejiang Gongshang University, Hangzhou}
  \country{China}
}
\affiliation{
  \institution{Zhejiang Key Laboratory of Big Data and Future E-Commerce Technology, Hangzhou}
  \country{China}
}

\author{Qiutao Song}
\affiliation{
  \institution{School of Computer Science and Technology, Zhejiang Gongshang University, Hangzhou}
  \country{China}
}

\author{Ziyi Wang}
\affiliation{
  \institution{School of Computer Science and Technology, Zhejiang Gongshang University, Hangzhou}
  \country{China}
}

\author{Linyang He}
\affiliation{
  \institution{Jianpei Technology Co., Ltd., Hangzhou}
  \country{China}
}

\author{Weiping Ding}
\affiliation{
  \institution{School of Artificial Intelligence and Computer Science, Nantong University, Nantong}
  \country{China}
}

\author{Mahmoud Hassaballah}
\affiliation{
  \institution{Department of Computer Science, College of Computer Engineering and Sciences, Prince Sattam Bin Abdulaziz University, Al Kharj}
  \country{Saudi Arabia}
}
\affiliation{
  \institution{Department of Computer Science, Faculty of Computers and Information, Qena University, Qena}
  \country{Egypt}
}

\author{Karen Egiazarian}
\affiliation{
  \institution{Department of Computing Sciences, Faculty of Information Technology and Communication Sciences, Tampere University, Tampere}
  \country{Finland}
}

\author{Wei-fa Yang}
\affiliation{
  \institution{Division of Oral and Maxillofacial Surgery, Faculty of Dentistry, The University of Hong Kong, Hong Kong}
  \country{China}
}

\author{Leszek Rutkowski}
\affiliation{
  \institution{Systems Research Institute of the Polish Academy of Sciences, Warsaw}
  \country{Poland}
}
\affiliation{
  \institution{AGH University of Krakow, Kraków}
  \country{Poland}
}
\affiliation{
  \institution{SAN University, Łódź}
  \country{Poland}
}

\renewcommand{\shortauthors}{Pengcheng Xue, Yan Tian et al.}

\begin{abstract}
Text-driven 3D scene editing has attracted considerable interest due to its convenience and user-friendliness. However, methods that rely on implicit 3D representations, such as Neural Radiance Fields (NeRF), while effective in rendering complex scenes, are hindered by slow processing speeds and limited control over specific regions of the scene. Moreover, existing approaches, including Instruct-NeRF2NeRF and GaussianEditor, which utilize multi-view editing strategies, frequently produce inconsistent results across different views when executing text instructions. This inconsistency can adversely affect the overall performance of the model, complicating the task of balancing the consistency of editing results with editing efficiency. To address these challenges, we propose a novel method termed Single-View to 3D Object Editing via Gaussian Splatting (SVGS), which is a single-view text-driven editing technique based on 3D Gaussian Splatting (3DGS). Specifically, in response to text instructions, we introduce a single-view editing strategy grounded in multi-view diffusion models, which reconstructs 3D scenes by leveraging only those views that yield consistent editing results. Additionally, we employ sparse 3D Gaussian Splatting as the 3D representation, which significantly enhances editing efficiency. We conducted a comparative analysis of SVGS against existing baseline methods across various scene settings, and the results indicate that SVGS outperforms its counterparts in both editing capability and processing speed, representing a significant advancement in 3D editing technology. For further details, please visit our project page at: \href{https://amateurc.github.io/svgs.github.io/}{https://amateurc.github.io/svgs.github.io/}.
\end{abstract}




\maketitle

\section{Introduction}\label{sec:introduction}
3D dimensional editing has emerged as a prominent area of research within the interdisciplinary fields of computer vision and graphics. A fundamental challenge in this domain is the effective reconstruction of three-dimensional objects from limited two-dimensional visual data, as well as the execution of precise modifications to these objects. Neural Radiance Fields (NeRF) \cite{mildenhall2021nerf} represent a sophisticated approach for reconstructing realistic representations of real-world 3D scenes. The primary advantage of NeRF lies in its capacity to simulate the behavior of light within three-dimensional space through intricate algorithms, thereby producing highly realistic visual effects. However, despite the substantial advancements in visual realism achieved by NeRF, it continues to encounter limitations regarding content editing. Most NeRF models utilize Multilayer Perceptrons (MLP) or Spherical Harmonics for the implicit encoding of 3D scenes, a technique that permits the observation of the shape and appearance of the scene only after rendering. Consequently, in comparison to explicit scenes, the potential for content editing within NeRF is significantly restricted.

Previous research has investigated neural field editing techniques, including the modification of texture and color of objects through alterations in latent codes \cite{schwarz2020graf,liu2021editing,jang2021codenerf,wang2022clip}, the creation of individual radiance fields for each object to facilitate the editing of specific items \cite{niemeyer2021giraffe,kundu2022panoptic}, and the integration of semantic information into neural fields to enable semantic-based editing operations \cite{wang2022clip,jain2022zero}. Nonetheless, these editing processes often necessitate substantial user input, such as a considerable amount of image data for the construction of radiance fields. Furthermore, these methodologies do not adequately address the requirements for creative editing, exemplified by the transformation of a pickup truck into a sedan. Consequently, there is a pressing need for further research and development of user-friendly and precise three-dimensional editing methods to enhance the re-creation of existing three-dimensional assets.

\begin{figure*}
    \centering
    \includegraphics[width=\textwidth]{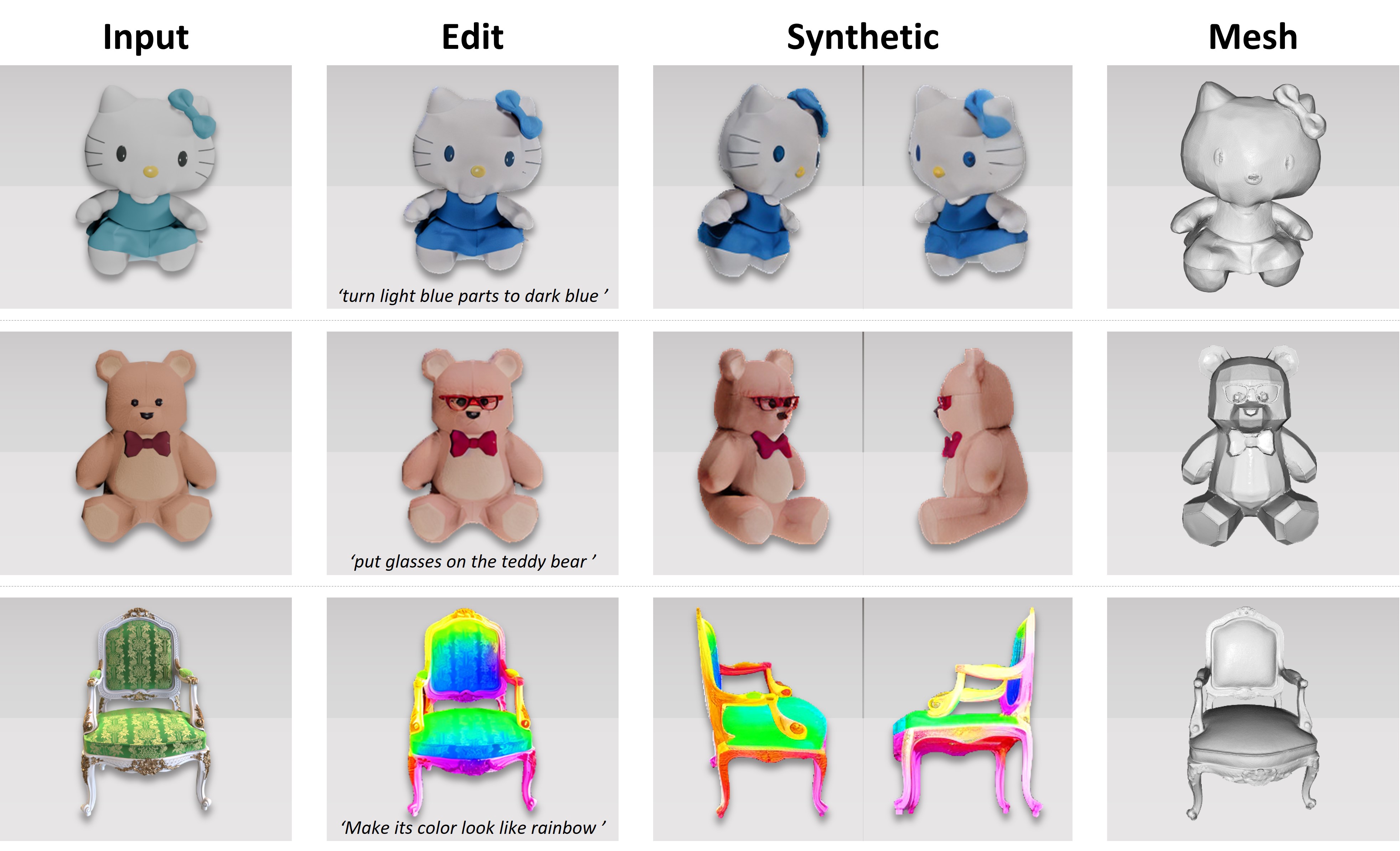}
    \caption{\textbf{Results of 3D Object Editing with SVGS}. This study demonstrates that text-driven 3D object editing can be accomplished through the modification of a single image, allowing for the generation of consistent multi-view images based on the edited outcomes. Notably, our algorithm facilitates precise alterations, such as changing a light blue hue to a dark blue hue, without impacting other regions of the object. Furthermore, the edited model can be exported as a high-quality mesh.}
    \label{fig:show_result}
\end{figure*}

\par\indent The recent advent of diffusion models \cite{ho2020denoising} has significantly enhanced the efficacy of generating image content from textual prompts. Notably, frameworks such as DreamFusion \cite{poole2022dreamfusion} and Magic3D \cite{lin2023magic3d} have integrated pretrained diffusion models with 3D representations through score distillation sampling. While these standard single-image 3D generation frameworks have achieved remarkable success in generation tasks, directly adapting them for editing presents unique challenges that standard pipelines cannot address. Existing multi-view editing approaches \cite{haque2023instruct,chen2024gaussianeditor,wang2024gaussianeditor,dong2024vica,zhuang2024tip,wu2024gaussctrl,tian2023survey} often struggle to reconcile the consistency of editing outcomes with operational efficiency. Specifically, a naive application of generative backbones creates a fundamental conflict between creative modification and geometric preservation.

\par\indent In response to these challenges, we propose SVGS, an advanced framework specifically architected to facilitate flexible and efficient fine-grained editing. We observe that standard 2D editors (e.g., IP2P \cite{brooks2023instructpix2pix}), when utilized directly as in prior works \cite{haque2023instruct,zhuang2024tip}, suffer from severe over-editing, where background elements or non-target regions are destructively hallucinated. Furthermore, standard sparse-view reconstruction via 3DGS typically leads to geometry collapse and floating artifacts when the generative prior from multi-view diffusion models \cite{long2024wonder3d} contains inconsistencies. Therefore, we argue that architectural innovation is required not merely in combining these models, but in designing specific constraint mechanisms to bridge the gap between 2D creativity and 3D consistency.

\par\indent Our framework addresses these intrinsic conflicts through a novel pipeline design featuring two key architectural constraints. (1) We introduce a Relevance-Aware Editing mechanism to tame the hallucination of 2D diffusion models. By incorporating both null and editing instructions into the U-Net of IP2P and analyzing the differential noise response, we explicitly predict the semantic relevance of each pixel, ensuring that modifications are strictly confined to the target area (e.g., changing only the ``light blue'' parts as shown in Fig.~\ref{fig:show_result}). (2) To resolve the sparse-view ambiguity in 3D reconstruction, we propose a Structural Prior Initialization method. Instead of relying on random initialization which leads to local minima, we leverage visual hulls derived from the generated views to strictly bound the 3D Gaussian optimization space. (3) Finally, we enhance the pipeline with Gaussian depth regularization to further refine the geometry. This constraint-based design ensures that SVGS achieves superior editing fidelity compared to standard generation pipelines.

\par\indent The contributions of this study can be summarized as follows:
\begin{itemize}
    \item[$\bullet$]We propose SVGS, a 3D object editing framework that introduces specific architectural constraints to standard generative pipelines. It effectively resolves the trade-off between editing creativity and geometric consistency, enabling precise text-driven modifications from a single view.
    \item[$\bullet$]We design a novel correlation editing strategy (Relevance-Aware mechanism) that extracts semantic masks from diffusion noise differences. This effectively prevents the over-editing problem inherent in standard IP2P implementations.
    \item[$\bullet$]We propose a structural prior initialization method that optimizes 3D Gaussians by utilizing explicit structural priors derived from highly sparse views. This architectural choice effectively mitigates the floating artifacts and geometry collapse that arise during standard sparse view reconstruction.
\end{itemize}

We conducted a thorough evaluation of the SVGS across multiple scenarios. In comparison to existing methodologies, SVGS consistently exhibits superior performance regarding editing quality, efficiency, and user satisfaction.

This paper is organized as follows. Section {\ref{related work}} presents works related to our approach. Section {\ref{preliminaries}} briefly reviews the strategies of original 3DGS and Multi-View Diffusion Models. Section {\ref{Method}} provides a detailed introduction to the proposed framework and its specific implementation methods. Finally, we evaluate our SVGS pipeline in section {\ref{experiments}}.

\section{Related Work}
	\label{related work}
	\subsection{3D Representations}
        \label{3D Representations}
	Mildenhall et al. \cite{mildenhall2021nerf} introduced the NeRF technique, which synthesizes novel views of complex static scenes from a set of well-posed input images. The fundamental concept involves modeling the continuous radiance field of the scene using a MLP, and subsequently synthesizing images through differentiable volume rendering while backpropagating photometric errors. NeRF has garnered significant attention within the field of computer vision due to its simplicity, user-friendliness, and exceptional performance. However, a notable limitation is that the optimization process for NeRF is exceedingly time-consuming. Recently, 3DGS \cite{kerbl3Dgaussians} has emerged as an alternative three-dimensional representation to NeRF, demonstrating remarkable quality and speed in 3D reconstruction tasks, and attracting considerable research interest in the generative domain \cite{chen2024text,tang2024lgm,tang2023dreamgaussian}. In this study, we are the first to apply 3DGS to single-view-based 3D editing tasks, with the objective of achieving rapid and efficient 3D editing\cite{zhao2026innovative}.
	
	\subsection{3D Editing}
        \label{3D Editing}
	Editing 3D scenes encompasses the manipulation of model geometry and color within a scene, guided by user input. A substantial amount of research has been conducted on the methodologies for editing 3D scenes. For instance, Neural Sparse Voxel Fields (NSVF) \cite{liu2020neural} employs a sparse voxel octree structure to encapsulate both geometric and appearance information of a scene. This approach not only enhances the computational efficiency of the NeRF \cite{mildenhall2021nerf} but also facilitates straightforward scene editing through voxel manipulation. However, NSVF is limited in its ability to manage complex backgrounds. To further augment editing capabilities, Generative Auto-Regressive Flows (GARF) \cite{schwarz2020graf} integrates Generative Adversarial Networks (GANs) \cite{goodfellow2014generative} with NeRF, enabling the generation of images with controllable scenes. Object-NeRF \cite{yang2021learning} introduces an innovative dual-path architecture that encodes the geometric and appearance information of a scene, allowing for individualized encoding of each object and facilitating object movements and rotations. Nonetheless, these methodologies may encounter challenges in decoupling factors of variation when inherent biases are present in the data. To achieve improved decoupling, Code-NeRF \cite{jang2021codenerf} proposes a joint learning framework that utilizes distinct shape and texture embeddings. This design allows for explicit control and editing of shape and texture, thereby providing enhanced precision in the modification of object characteristics without the complications of coupling. Giraffe \cite{niemeyer2021giraffe} and PNF \cite{kundu2022panoptic} tackle this issue by establishing separate neural radiance fields for each object; however, these approaches may lack user-friendliness in the editing process. The groundbreaking work EditNeRF \cite{liu2021editing} trains category-level NeRFs, also referred to as Conditional Radiation Fields, on shape categories to translate rough 2D user sketches into 3D space for the modification of color or shape in localized regions. Nevertheless, this method is constrained in its shape processing capabilities, as it only permits the addition or removal of localized object parts. In contrast, CLIP-NeRF \cite{wang2022clip} facilitates global deformations in shape manipulation and enhances editing capabilities by incorporating the CLIP model into the workflow. This integration allows for text and image-based editing, rendering the process more accessible, particularly for novice users. While the application of CLIP to optimize NeRF parameters for editing or generating images that correspond to text prompts \cite{jain2022zero,wang2022clip} shows potential, these methods often fall short in achieving precise selective editing of specific regions within a scene. Haque et al. \cite{haque2023instruct} employed IP2P \cite{brooks2023instructpix2pix} to facilitate text-based interactive editing. Although this approach yielded convincing results, it was hindered by low editing efficiency. Recent research has focused on improving editing efficiency, with numerous works \cite{chen2024gaussianeditor,tang2023dreamgaussian,wang2024gaussianeditor,zhuang2024tip,palandra2024gsedit} have applied 3DGS \cite{kerbl3Dgaussians} to 3D editing tasks. 
    
    However, these methods still struggle to effectively address the issue of inconsistency in editing outcomes. Specifically, during multi-view editing, the effects of editing across different perspectives often lack consistency, resulting in discrepancies between certain views. This inconsistency can lead to distortions in the overall 3D reconstruction, failing to satisfy users' demands for high-precision and continuous editing. The challenge is further magnified when dealing with complex objects or scenes, where viewpoint differences become more pronounced. Consequently, achieving a balance between editing efficiency and ensuring coherence and consistency across all viewpoints continues to be a major technical challenge that needs to be addressed. In contrast, our proposed method integrates 3DGS \cite{kerbl3Dgaussians} with a multi-view diffusion model, markedly enhancing the quality, consistency, and efficiency of editing results. We demonstrate the advantages of single-view-based editing, illustrating that the entire training and editing process is both straightforward and efficient.




\section{Preliminaries}
	\label{preliminaries}

	\subsection{3D Gaussian Splatting}
3DGS \cite{kerbl3Dgaussians} has rapidly gained considerable attention for its superior rendering quality and efficiency. This technique utilizes point clouds to represent explicit three-dimensional scenes and employs Gaussian functions to characterize the structure of these scenes. Each 3D Gaussian is defined by a position \(\mu\), a full 3D covariance matrix \(\boldsymbol{\Sigma}\), opacity \(\alpha\), and color, which is represented through spherical harmonics (\(\mathbf{SH}\)). To render these 3D Gaussians, they are projected onto 2D Gaussians. We adhere to the projection method proposed by Zwicker et al. \cite{zwicker2001ewa} to derive the view-space covariance matrix \(\boldsymbol{\Sigma'}\):
\begin{equation}
\mathbf{\Sigma'} = \mathbf{JW{\Sigma}W^{T}J^{T}},
\end{equation}
where \(\mathbf{W}\) represents the view transformation, while \(\mathbf{J}\) denotes the Jacobian matrix associated with the affine approximation of the projection transformation.

The physical interpretation of the covariance matrix holds true only when the matrix is positive semi-definite, which prevents its direct optimization for representing a scene's  radiance field. Nevertheless, differentiable optimization can be attained by decomposing the covariance matrix \(\mathbf{\Sigma}\) into a rotation matrix \(\mathbf{R}\) and a scaling matrix \(\mathbf{S}\):
\begin{equation}
\mathbf{\Sigma} = \mathbf{RSS^TR^T}.
\end{equation}
The calculation of gradient flow is explained detail in \cite{kerbl3Dgaussians}.After projection, the 2D Gaussians are organized based on depth information. Utilizing a point-based approach, the color \(\mathbf{C}\) is determined by blending the \(\mathbf{N}\) ordered points that coincide with the pixel:
\begin{equation}
\mathbf{C} = \sum_{i \in N} c_i \alpha_i \prod_{j=1}^{i-1} (1 - \alpha_j).
\end{equation}
In this context, \(c_i\) and \(\alpha_i\) denote the color and density of a specific point, respectively. These parameters are derived from a Gaussian distribution characterized by the covariance matrix \(\boldsymbol{\Sigma}\), and are adjusted by the optimizable per-point opacity and the coefficients of spherical harmonics associated with color.	
\subsection{Multi-View Diffusion Models}
Diffusion models \cite{sohl2015deep,ho2020denoising,LDM}, similar to other generative models, are capable of generating target data samples from noise that is sampled from a standard normal distribution. Recently, these models have gained popularity for their ability to generate high-quality synthetic images. The architecture of these models is centered around two fundamental processes. The first process is the forward process, in which a sample \(x_0\) drawn from a data distribution \(p(x)\) undergoes a gradual degradation of its structure through the addition of random noise over \(t\) steps. The stochastic process \(\{ X_t \}_{t=0}^T,\) is mathematically represented as:
\begin{equation}
X_t = \sqrt{\bar{\alpha}_t} X_0 + \sqrt{1 - \bar{\alpha}_t} \epsilon, \quad \text{where} \quad \epsilon \sim \mathcal{N}(0, I).
\end{equation}
Here \( t \in [0, T] \) represents the noise level, and \( \bar{\alpha}_{1:T} \in (0, 1]^T \) denotes a decreasing sequence. The second process is a reverse operation that gradually eliminates the noise while reintroducing structure into the data sample. This process entails the recovery of \( x_{t-1} \) from \( x_t \) by subtracting the predicted added random noise \(\epsilon\). The reverse process is parameterized by a conditional neural network \( \epsilon_{\theta} \), which is trained to predict the noise \( \epsilon \) based on the following simplified objective \cite{ho2020denoising}:
\begin{equation}
\mathbb{E}_{x \sim p(x), \epsilon \sim \mathcal{N}(0, I), t \sim \mathcal{T}} \left[ \| \epsilon_\theta(x_t, t, c) - \epsilon \|_2^2 \right],
\end{equation}
where \( \mathbf{c} \) represents a condition (such as text, image, etc.) that facilitates the regulation of the generation process, while \( \mathcal{T} \) represents a set comprising a selection of timesteps.

To mitigate the multi-view consistency issue in 2D lifting approaches, a typical solution involves enhancing their viewpoint-aware capabilities. For instance, Dreamfusion\cite{poole2022dreamfusion} incorporated viewpoint descriptions into text conditioning. However, we observe that even a perfectly camera-conditioned model remains insufficient to resolve this problem, as content discrepancies across different perspectives may still persist. Through these observations, we identify the critical importance of directly training a multi-view diffusion model, leveraging 3D-rendered datasets of static scenes with precise camera parameters. Our approach employs 3D datasets to render consistent multi-view images for supervised training of the diffusion model. Formally, given a set of noisy images $x_t \in \mathbb{R}^{F \times H \times W \times C}
$, a text prompt as condition $y$, and a set of extrinsic camera parameters $c \in \mathbb{R}^{F \times 16}
$, we train the multi-view diffusion model to generate an image set $x_0 \in \mathbb{R}^{F \times H \times W \times C}
$ representing the same scene from F distinct viewpoints. Once trained, the model can serve as a multi-view prior.

\begin{figure*}
    \centering
    \includegraphics[width=\linewidth]{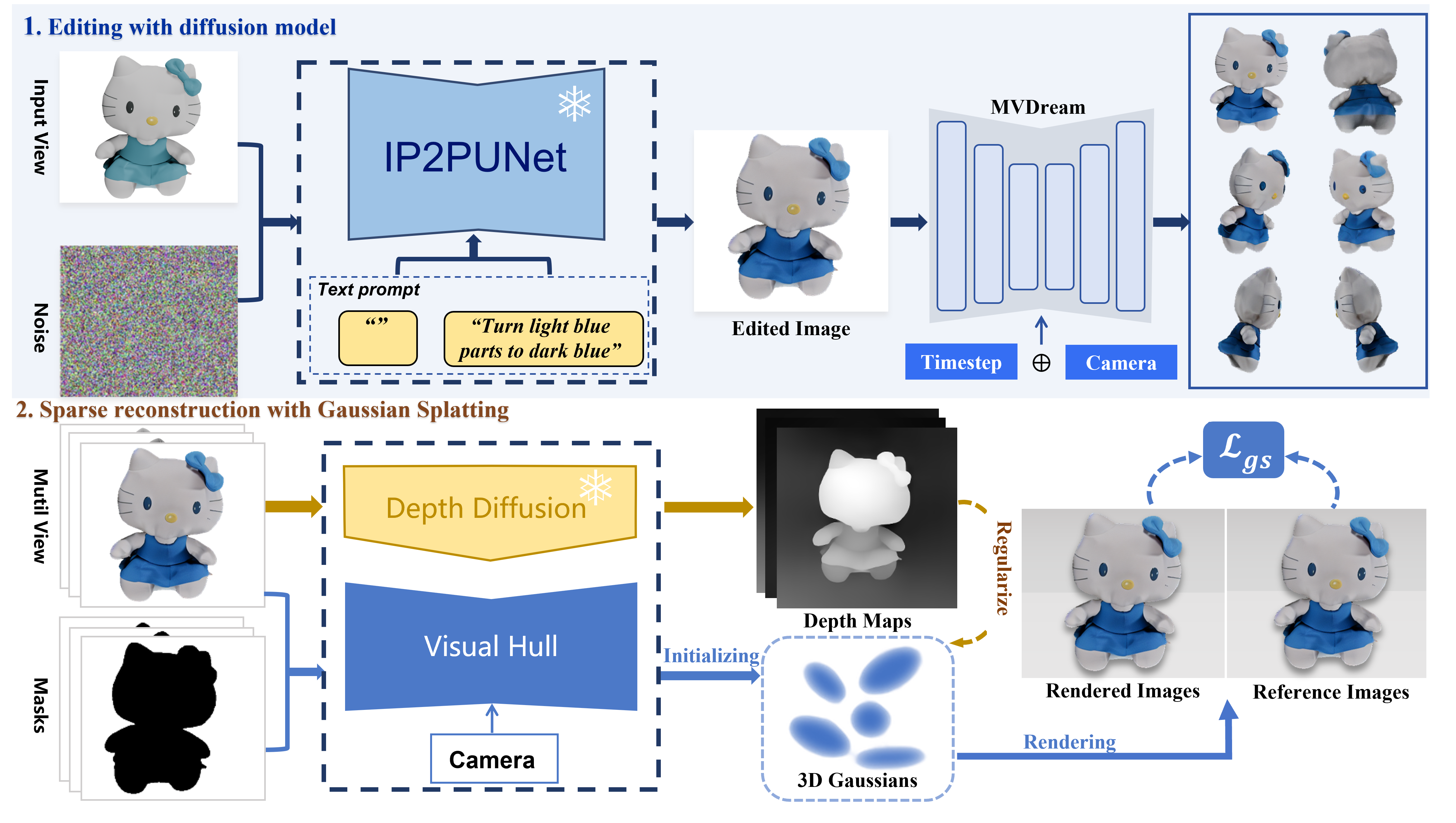}
    \caption{\textbf{The Framework of SVGS.} 1) Initially, a single-view image is modified using a correlation editing strategy based on IP2P. 2) Following the editing process, a multi-view diffusion model is employed to generate multi-view images consistent with the modifications. 3) For sparse reconstruction using 3DGS, a visual hull is constructed from camera parameters and masked images to initialize the 3D Gaussians, which are subsequently refined using the loss function $\mathcal{L}_{gs}$. 4) Within the depth regularization module, depth maps are rendered for the input views, and the loss is computed against pre-generated monocular depth maps. The resulting Gaussian field facilitates efficient, high-quality novel-view synthesis.}
    \label{fig:pipeline}
\end{figure*}
	
	\section{Method}
	\label{Method}

\subsection{Overview}
Our proposed method, referred to as SVGS, and illustrated in Fig. \ref{fig:pipeline}, aims to leverage a generative 2D diffusion model \cite{brooks2023instructpix2pix,LDM} for the editing of single-view images, while integrating 3DGS technology \cite{kerbl3Dgaussians} to facilitate 3D editing tasks. However, unlike standard single-image generation frameworks that synthesize content from scratch, an editing pipeline must resolve the intrinsic conflict between creative modification and geometric preservation. A naive combination of these components typically fails due to two primary challenges.

Firstly, existing diffusion-based image editors\cite{tian2026DM-CFO} lack automated mechanisms for identifying editing regions. Although IP2P \cite{brooks2023instructpix2pix} represents the most advanced approach, we observe that it frequently results in excessive alterations (over-editing) when applied globally, destroying the original identity of the object. Secondly, regarding the 3D representation, we adopt the latest advancements in 3DGS \cite{kerbl3Dgaussians} for its efficiency. Nonetheless, directly training 3DGS on the sparse views generated by multi-view diffusion models is prone to overfitting and geometric collapse (local minima), as standard Structure-from-Motion (SfM) initialization is inapplicable in this synthetic sparse setting.

To address these architectural gaps, we propose SVGS as a constraint-based pipeline. In Section \textcolor{red}{\ref{sec:Relevance Editing}}, we delineate a Relevance-Aware Editing technique. Instead of relying on manual masks, this module predicts the relevance of each pixel to the editing instruction by analyzing diffusion noise differences, thereby strictly constraining the editing area. In Section \textcolor{red}{\ref{sec:Structural Prior Initialization}}, we present a Structural Prior Initialization method. To prevent geometric collapse in sparse-view reconstruction, we utilize visual hulls derived from the generated views to provide a robust geometric initialization. Finally, in Section \textcolor{red}{\ref{sec:Depth Regularization for Gaussians}}, we introduce a Gaussian depth regularization technique aimed at producing high-quality, artifact-free 3D reconstructions.

\subsection{Relevance-Aware Editing}
\label{sec:Relevance Editing}
Current diffusion-based image editing techniques typically lack effective mechanisms for the automatic localization of the editing region. These methods either necessitate user-provided masks \cite{lugmayr2022repaint}, depend on global information retained in the noise input \cite{meng2021sdedit}, or directly apply the denoiser to the input \cite{brooks2023instructpix2pix}. However, each of these approaches is prone to excessive alterations, often modifying background details that should remain preserved \cite{brooks2023instructpix2pix,couairon2022diffedit}. To facilitate precise text-based editing, it is imperative to delineate the editing region as specified by the textual instructions.

\begin{figure}[t]
\centering
\includegraphics[width=\linewidth]{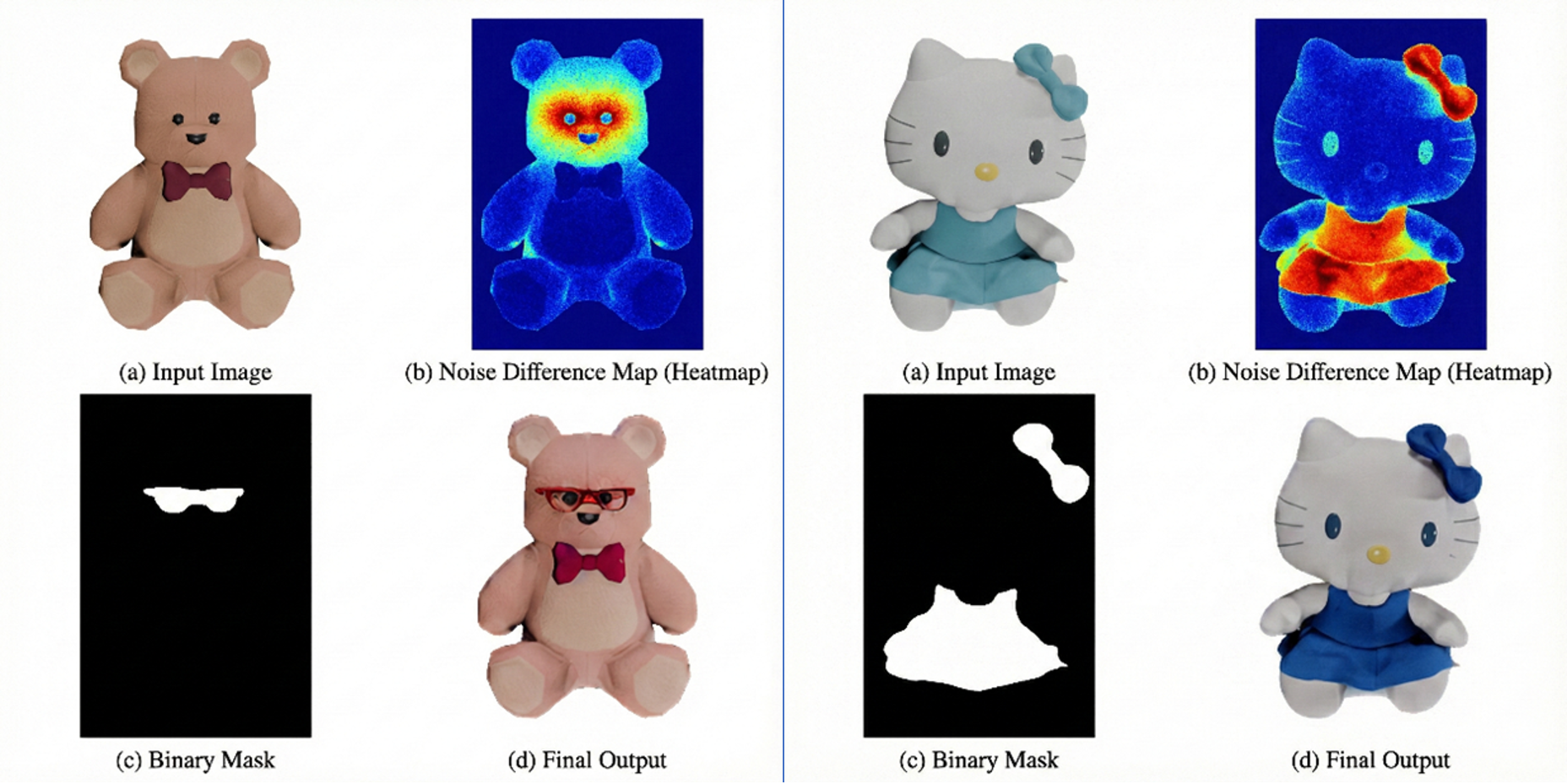}
\caption{\textbf{Visualization of the Relevance-Aware Editing Mechanism.} We visualize the intermediate steps of our algorithm. The \textbf{Noise Difference Heatmap (b)}, derived from Eq. (\ref{eq:relevance_map}), effectively acts as a semantic attention mechanism. It highlights the target region (e.g., the eyes for ``put glasses on'') where the text instruction exerts significant influence, while suppressing the background. This is thresholded into a binary mask (c) to ensure the final output strictly preserves the unedited content.}
\label{fig:relevance_viz}
\end{figure}

To achieve this, we introduce a Relevance-Aware Editing mechanism grounded in the IP2P framework \cite{brooks2023instructpix2pix}. The core intuition is that the diffusion model's noise prediction will vary significantly in regions pertinent to the editing instruction compared to a null instruction, while remaining consistent in irrelevant background areas. As illustrated in Fig. \ref{fig:relevance_viz}, we visualize this process: the differential noise response forms a clear heatmap centered on the target object parts.

Specifically, given a single-view image $I$ and a text instruction $C_T$, we employ the IP2P framework to assess the relevance of each pixel. Initially, for the image $I$, the diffusion process introduces random noise to the encoded latent representation $z = \mathcal{E}(I)$ until the timestep $t_r$, resulting in a noisy latent representation:
\begin{equation}
z_{t_r} = \sqrt{\bar{\alpha}_{t_r}} \mathcal{E}(I) + \sqrt{1 - \bar{\alpha}_{t_{r}}} \epsilon, \quad
\end{equation}
where \(\epsilon \sim \mathcal{N}(0, 1)\) represents the random noise, and \(\alpha_t\) denotes the noise scheduling factor. We then use the U-Net \(\epsilon_\theta\) to generate two distinct predictions: i) the predicted noise conditioned on the text instruction \(\epsilon_{I,T}(z_{t_{\text{r}}}) = \epsilon_\theta(z_{t_{\text{r}}}, t_{\text{r}}, I, C_T)\); and ii) the predicted noise conditioned on a null instruction, represented as \(\epsilon_I(z_{t_{\text{r}}}) = \epsilon_\theta(z_{t_{\text{r}}}, t_{\text{r}}, I, \emptyset) \).

To measure the semantic relevance of each pixel, we compute the absolute difference between these two predictions:
\begin{equation}
R_{x,I,T} = |\epsilon_{I,T}(z_{t_{\text{r}}}) - \epsilon_I(z_{t_{\text{r}}})|.
\label{eq:relevance_map}
\end{equation}
As shown in Fig. \ref{fig:relevance_viz} (b), this difference map \(R_{x}\) effectively acts as an attention mechanism. We normalize this map and apply a threshold to generate a binary mask \(M\), which is then used to blend the edited latent with the original latent, ensuring that background regions remain strictly unaltered.

Algorithm \ref{alg:relevance_editing} outlines the complete procedure. By assessing and localizing the relevance of each pixel, this methodology significantly minimizes unnecessary global modifications and avoids the negative effects associated with over-editing commonly observed in traditional methods.

\begin{algorithm}[H]
\caption{Relevance-Aware Editing with IP2P}
\label{alg:relevance_editing}
\KwIn{$I$ (image), $C_T$ (text instruction), $t_r$ (timestep), $\mathcal{E}$ (encoder), $\epsilon_\theta$ (noise prediction model)}
\KwOut{$M$ (editing mask)}

$z \gets \mathcal{E}(I)$ \tcp*[r]{Encode image}
$\epsilon \sim \mathcal{N}(0, 1)$ \tcp*[r]{Sample random noise}
$\bar{\alpha}_{t_r} \gets \prod_{t=1}^{t_r} \alpha_t$ \tcp*[r]{Compute noise scheduling factor}

$z_{t_r} \gets \sqrt{\bar{\alpha}_{t_r}} \cdot z + \sqrt{1 - \bar{\alpha}_{t_r}} \cdot \epsilon$ \tcp*[r]{Add noise}

$\epsilon_{I,T} \gets \epsilon_\theta(z_{t_r}, t_r, I, C_T)$ \tcp*[r]{Text-conditioned prediction}
$\epsilon_{I} \gets \epsilon_\theta(z_{t_r}, t_r, I, \emptyset)$ \tcp*[r]{Null-conditioned prediction}

$R_x \gets |\epsilon_{I,T} - \epsilon_{I}|$ \tcp*[r]{Compute differential noise}
$R_x \gets \text{Normalize}(R_x)$ \tcp*[r]{Normalize to [0, 1]}
$M \gets \{x \mid R_x[x] > \tau \}$ \tcp*[r]{Thresholding to binary mask}

\Return $M$
\end{algorithm}

\begin{figure*}
\centering
\includegraphics[width=\linewidth]{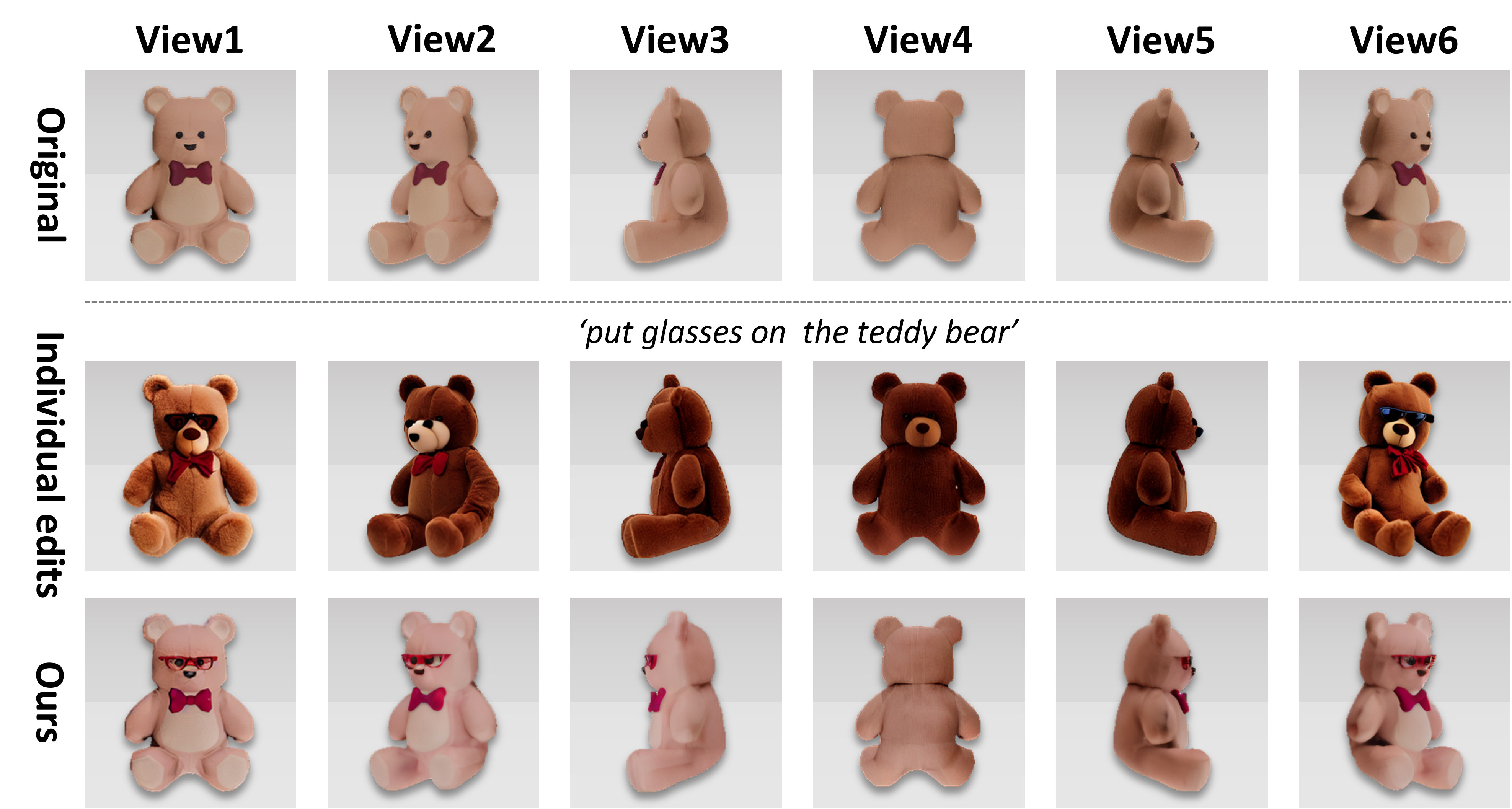}
\captionof{figure}{\textbf{Consistency Experiments.} The top row presents six images sourced from the original dataset, whereas the middle row illustrates the outcomes of independently editing each view utilizing the IP2P method. The results of these edits exhibit significant inconsistency and do not correspond effectively with the provided textual instructions. In contrast, our proposed method produces consistent and well-edited multi-view images, as demonstrated in the bottom row.}
\label{fig:conherent}
\end{figure*}
\subsection{Structural Prior Initialization}
\label{sec:Structural Prior Initialization}
To obtain consistent editing results and more view information, we generated multi-view images from six different viewpoints using a pre-trained multi-view diffusion model, and utilized them for 3D Gaussian (3DGS) reconstruction \cite{kerbl3Dgaussians}. The generated results are illustrated in Fig. \. However, the sparse views provided by these six images yield very limited three-dimensional information, particularly because the critical SfM points necessary for the initialization of 3DGS \cite{kerbl3Dgaussians} are frequently absent. Consequently, we propose a structure-prior-based initialization method that fully exploits the structural information derived from the limited views to achieve a satisfactory object contour.
\\\indent In this study, we employ contour images and object masks to construct a visual hull, which acts as the geometric framework for initializing the 3D Gaussian distribution. In contrast to SfM points, which offer limited information, the visual hull supplies a more comprehensive structural prior, thereby facilitating the elimination of implausible Gaussian distributions and promoting multi-view consistency. The mask images required for the creation of the visual hull can be readily acquired through segmentation models \cite{kirillov2023sam}.\par
To initialize the 3D points, we first establish a voxel space and subsequently generate 3D points \( P_i(x_i, y_i, z_i) \) randomly within this space, which we denote as the set \(\mathcal{P}\). For each 3D point \( P_i \in \mathcal{P} \), we project it onto the plane corresponding to each view \( k \) utilizing the projection matrix \( P_k \), and we assess whether it lies within the silhouette
\begin{equation}
P_k = K_k [R_k \mid t_k],
\end{equation}
where \( K_k \) represents the intrinsic matrix of the camera, while \( R_k \) and \( t_k \) denote the rotation matrix and translation vector, respectively. A point \( P_i \) is preserved if it lies within the projection silhouette of all views
\begin{equation}
P_i \in \bigcap_{k=1}^n S_k,
\end{equation}
where \( S_k \) represents the binary silhouette image corresponding to the \( k \)-th view. Subsequently, for the retained 3D points, their color values are determined based on their projected positions in the reference images through bilinear interpolation. The average color value across all views is then computed
\begin{equation}
C_i = \frac{1}{n} \sum_{k=1}^n I_k(u_i, v_i),
\end{equation}
In this context, \( I_k \) denotes the reference image corresponding to the \( k \)-th view, while \((u_i, v_i)\) represents the projected coordinates of the point \( P_i \) on the image \( I_k \). The color value is derived using bilinear interpolation. Subsequently, these 3D points are converted into 3D Gaussians.\par
We optimize the initial \( \mathcal{G}_i \) using the 3DGS method as described by Kerbl et al. \cite{kerbl3Dgaussians}. The initial optimization process for \( \mathcal{G}_i \) incorporates both color and mask losses. The color loss is defined as a combination of L1 loss and D-SSIM loss
\begin{equation}
\mathcal{L}_1 = \| I - I^{\text{ref}} \|_1, \quad \mathcal{L}_{\text{D-SSIM}} = 1 - \text{SSIM}(I, I^{\text{ref}}),
\end{equation}
where $I$ is the rendered image and $I^{\text{ref}}$ is the corresponding reference image. A binary cross-entropy (BCE) loss function is utilized to compute the mask loss
\begin{equation}
\mathcal{L}_m = m^{\text{ref}}log((1-m)/m) - \log (1 - m),
\end{equation}
where $m$ denotes the object mask.
\\\indent The structure-prior-based initialization method presents several significant advantages. By utilizing contour images and object masks to construct a visual hull, this approach offers a more comprehensive source of structural information compared to traditional SfM points, which are frequently sparse and incomplete. Consequently, this leads to more accurate and consistent 3D reconstructions, as the visual hull provides a dependable geometric framework for initializing the 3D Gaussian distribution. Furthermore, the method aids in the elimination of unreasonable Gaussian placements and ensures multi-view consistency, which is essential for achieving high-quality 3D reconstructions. The accessibility of mask images through segmentation models~\cite{tian20223d,tian2024rgb,tian2023revised} further enhances the practicality of this method, facilitating a more efficient and robust initialization process in scenarios with limited views. Ultimately, this approach enhances both the accuracy of object contours and the overall fidelity of the reconstructed 3D models.

\subsection{Depth Regularization for Gaussians}
\label{sec:Depth Regularization for Gaussians}
The implementation of a structure prior initialization strategy can substantially improve the performance of 3DGS in the context of novel view synthesis. While the overall fidelity of our reconstruction is deemed acceptable, certain challenges persist in poorly observed areas, occluded regions, and unobserved regions. In these instances, some surfaces within the rendered novel views exhibit semi-transparency, leading to the emergence of visible gaps.

To address these challenges, we draw inspiration from DnGaussian \cite{li2024dngaussian} and introduce Gaussian depth regularization. This regularization seeks to optimize the geometry by aligning the rendered accumulated depth with a pseudo-ground-truth depth map derived from a monocular estimator (e.g., ZoeDepth).

Formally, let $p$ denote a pixel in the image space $P$, and $\mathcal{N}(p)$ denote the set of ordered 3D Gaussians intersecting the ray passing through pixel $p$. The accumulated depth is calculated as:
\begin{equation}
\hat{D}(p) = \sum_{k \in \mathcal{N}(p)} d_k \, \alpha_k \prod_{j=1}^{k-1} (1 - \alpha_j),
\label{eq:depth_render}
\end{equation}
where $d_k$ is the depth of the $\mathcal{k}$-th Gaussian.

We then formulate the depth regularization loss $\mathcal{L}_{depth}$ by enforcing consistency between the rendered depth $\hat{D}$ and the monocular depth prior $D_{mono}$:
\begin{equation}
\mathcal{L}_{depth} = \frac{1}{|P|} \sum_{p \in P} \| \hat{D}(p) - D_{mono}(p) \|_2^2,
\label{eq:depth_loss}
\end{equation}
where $|P|$ is the total number of pixels.

The overall loss is a composite measure that integrates these components:
\begin{equation}
\mathcal{L}_{total} = (1 - \lambda_{\text{SSIM}}) \mathcal{L}_1 + \lambda_{\text{SSIM}} \mathcal{L}_{\text{D-SSIM}} + \lambda_m \mathcal{L}_m + \lambda_d \mathcal{L}_{depth},
\end{equation}
where $\lambda_{\text{SSIM}}$, $\lambda_{m}$, and $\lambda_{d}$ regulate the weights assigned to each term. By explicitly regularizing the depth, this method effectively addresses challenges associated with semi-transparent surfaces and visible gaps, significantly enhancing the geometric coherence of the novel view synthesis.

	\section{Experiments}
	\label{experiments}
\subsection{Experimental Setup}
\noindent\textbf{Implementation details}\hspace{0.5em}Our framework is built upon the PyTorch implementation of 3DGS \cite{kerbl3Dgaussians}. To harness the rich generative priors of large-scale models while maintaining computational efficiency, we leverage the pre-trained weights of IP2P \cite{brooks2023instructpix2pix} and MVDream \cite{shi2023mvdream} as frozen backbones. Crucially, our proposed Relevance-Aware Masking algorithm (Section \ref{sec:Relevance Editing}) and Structural Prior Initialization (Section \ref{sec:Structural Prior Initialization}) are implemented as plug-and-play modules that guide the optimization process without requiring fine-tuning of the diffusion U-Nets.

During the sparse reconstruction phase, the model was optimized for 10,000 iterations across all datasets. The loss function weights were set to $\gamma = 0.1$ and $\tau = 0.95$. Both input and rendered view resolutions were set to $512 \times 512$. In the editing phase, the relevance maps are computed dynamically during the diffusion inference steps. The final 3D refinement typically requires 1,000 to 3,000 iterations. The total processing time for the entire pipeline is approximately 20 minutes on a single RTX 4090 GPU, comprising roughly 15 minutes for the initial sparse reconstruction and 5 minutes for the text-driven editing stage. A detailed summary of all hyperparameters is provided in Table \ref{tab:hyperparams}.

\begin{table}[h]
\centering
\renewcommand{\arraystretch}{1.1} 
\caption{Hyperparameter settings for SVGS. Detailed parameters are listed to facilitate reproducibility.}
\begin{tabularx}{\linewidth}{X c} 
\toprule[1.5pt]
\textbf{Parameter} & \textbf{Value} \\
\midrule

\textbf{Hardware} & $3 \times$ RTX 4090, PyTorch \\
Render Resolution & $512 \times 512$ \\
\midrule[0.5pt]

\textbf{Stage 1: Sparse Recon.} & (10k Iterations) \\
Loss Params ($\gamma, \tau$) & $0.1, \ 0.95$ \\
\midrule[0.5pt]

\textbf{Stage 2: Text Editing} & (1k-3k Iterations) \\
Guidance Scales ($S_T, S_I$) & $7.5, \ 1.5$ \\
Relevance Threshold & 0.35 \\
\midrule[0.5pt]

\textbf{Regularization Weights} & \\
$\lambda_{SSIM}, \lambda_{m}, \lambda_{d}$ & $0.2, \ 0.1, \ 0.05$ \\

\bottomrule[1.5pt]
\end{tabularx}
\label{tab:hyperparams}
\end{table}

\vspace{0.5em}
\noindent\textbf{Mesh Extraction Settings.}\hspace{0.5em}To export high-quality meshes from the optimized 3D Gaussians, we employ the Gaussian-Opacity-Fields (GOF) extraction method \cite{gof}. Specifically, we extract the iso-surface using Marching Cubes with a grid resolution of $512^3$. An opacity threshold of $\tau_{opacity} = 0.3$ is applied to filter out low-density floaters. Post-processing includes Laplacian smoothing (5 iterations) to reduce high-frequency noise and mesh simplification (decimation) to reduce the face count to approximately 100,000 faces for efficient rendering.

\vspace{0.5em}
\noindent\textbf{Dataset.}\hspace{0.5em}To assess the efficacy and generalization capability of our proposed method, we conducted experiments utilizing two distinct datasets: the NeRF Blender synthetic dataset (Blender) \cite{gordon2023blended} and the Objaverse dataset \cite{deitke2024objaverse}. To ensure a robust evaluation, we expanded our test set to include 12 diverse objects (covering varying geometries, textures, and topologies) selected from these datasets. For each object, we systematically evaluated our method against a variety of text-based instructions ranging from texture modification to structural editing.

\vspace{0.5em}
\noindent\textbf{Baselines.}\hspace{0.5em}In the context of text-driven editing, we evaluate our methodology against the most recent editing techniques that utilize NeRF representation and GS representation, including (1) Instruct-N2N \cite{haque2023instruct}, which edits NeRF by iteratively updating the dataset with images generated by a 2D editing model; (2) GaussianEditor \cite{chen2024gaussianeditor}, which establishes a Hierarchical Generative Structure (HGS) framework for gradient-based optimization utilizing SDS loss, achieving semantic tracking through the application of two-dimensional semantic masks; and (3) ViCA-NeRF \cite{dong2024vica}, which employs two sources of regularization to explicitly propagate editing information across various views, thereby ensuring consistency across multiple perspectives.

\begin{figure*}
\centering
\includegraphics[width=\linewidth]{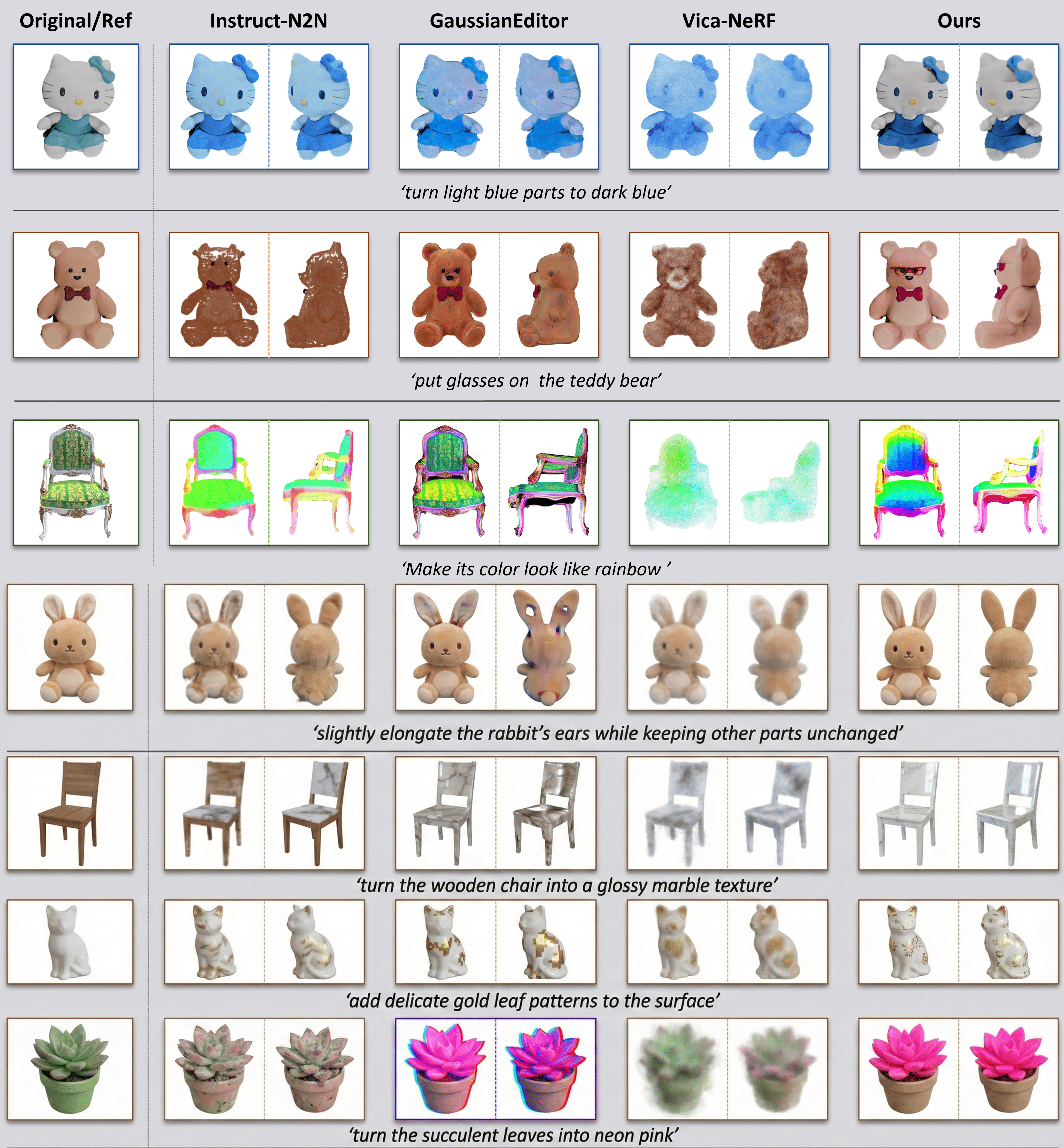}
\caption{Qualitative comparison of text-driven 3D object editing. The visual results are organized by methods (columns) and editing tasks (rows). The first column displays the original reference image, while the remaining columns present multi-view renderings from Instruct-N2N \cite{haque2023instruct}, GaussianEditor \cite{wang2024gaussianeditor}, ViCA-NeRF \cite{dong2024vica}, and our proposed SVGS. The seven rows correspond to distinct objects and editing instructions: (Row 1) color modification; (Row 2) accessory addition; (Row 3) style transfer; (Row 4) structural deformation (elongating rabbit ears); (Row 5) material transfer (wood to glossy marble); (Row 6) texture pattern addition (gold leaf); and (Row 7) localized part recoloring. Two views are rendered for each method to illustrate the appearance across different viewpoints.}
\label{fig:compare}
\end{figure*}

\subsection{Qualitative Evaluation}
Editing 3D objects is inherently a subjective endeavor; therefore, we primarily present a range of qualitative results and comparative analyses. As illustrated in Fig. \ref{fig:compare}, Instruct-N2N \cite{haque2023instruct} directly implements the editing outcomes derived from IP2P \cite{brooks2023instructpix2pix}, which results in excessive editing. Conversely, GaussianEditor \cite{chen2024gaussianeditor} employs semantic tracking to regulate the editing region; however, it encounters difficulties in achieving satisfactory results when provided with more nuanced editing instructions. In contrast, VICA-NeRF \cite{dong2024vica} tends to produce edits that lack sufficient detail. Notably, SVGS demonstrates superior performance compared to other methods in terms of both editing quality and controllability. The final outcomes, as depicted in Fig. \ref{fig:show_result}, can be extracted as meshes utilizing the gaussian-opacity-fields \cite{gof} technique. Our experiments encompass modifications to object features while maintaining original characteristics, as well as alterations to visual appearance.
\subsection{Quantitative Evaluation}
\noindent\textbf{User Study Protocol.}\hspace{0.5em}To rigorously evaluate the subjective quality of the editing results, we conducted a large-scale user study involving 120 participants. The participant pool consisted of domain experts, senior researchers, and 3D modeling practitioners with extensive background knowledge. The study followed a strictly double-blind protocol to prevent potential bias; all method names were anonymized, and the display order of the results was randomized for each sample. Participants were presented with the original input view, the text instruction, and the generated 3D renderings from different methods side-by-side. They were instructed to select the best result based on two key criteria: (1) Semantic Alignment (how accurately the edit reflects the textual instruction without missing parts) and (2) Perceptual Fidelity (geometric completeness, texture details, and multi-view consistency). Each participant evaluated a random subset of scenes from our dataset. The ``User Study \%'' reported in Table \ref{tab:comparison} represents the average preference rate. Statistical significance was verified using a paired t-test with $p < 0.05$.

\vspace{0.5em}
\noindent\textbf{Results Analysis.}\hspace{0.5em}In accordance with prior methodologies \cite{brooks2023instructpix2pix,haque2023instruct}, we employed CLIP directional similarity (CLIP-DS), user preference, and total processing time as evaluation metrics. As illustrated in Table~\ref{tab:comparison}, the Instruct-N2N model \cite{haque2023instruct} did not enhance the performance of the editing model IP2P \cite{brooks2023instructpix2pix}; rather, it merely optimized NeRF \cite{mildenhall2021nerf} through the IDU \cite{haque2023instruct} approach to improve the consistency of the editing outcomes, which consequently resulted in the longest processing time (51 min). GaussianEditor \cite{chen2024gaussianeditor} utilizes semantic tracking and a hierarchical Gaussian representation to delineate the editing area and mitigate the impact of edits on semantically unrelated regions, achieving a degree of success. Nonetheless, it remains inadequate for fine-grained editing tasks. ViCA-NeRF \cite{dong2024vica} seeks to enhance the quality of edited results by averaging latent codes to produce a singular final output; however, our experimental findings indicated minimal effectiveness. In contrast, SVGS not only demonstrated exceptional performance in user studies (60.32\%) but also excelled in CLIP directional similarity assessments (0.13), all while maintaining the lowest processing time (20 min).

\begin{table}[ht]
    \centering
    \setlength{\tabcolsep}{32pt}
        \caption{\textbf{Quantitative Evaluation}. SVGS outperforms in both user study evaluations and CLIP Directional Similarity \cite{CTIDS} metrics.}
    \begin{tabular}{lccc}
        \toprule
        Methods & CLIP-DS$\uparrow$ & Time$\downarrow$ & User Study$\uparrow$ \\
        \midrule
        Instruct-N2N [10] & 0.05 & 51 min & 15.24\%\\
        GaussianEditor [2] & 0.09 & 40 min & 17.72\%\\
        ViCA-NeRF [6] & 0.04 & 40 min & 6.72\%\\
        \midrule
        \rowcolor{gray!30}\textbf{Ours-SVGS} & \textbf{0.13} &  \textbf{20 min} & \textbf{60.32\%}\\
        \bottomrule
    \end{tabular}
    \label{tab:comparison}
\end{table}

\subsection{Ablation Studies}
\begin{figure}
\vspace{-12pt}
    \centering
    \includegraphics[width=\linewidth]{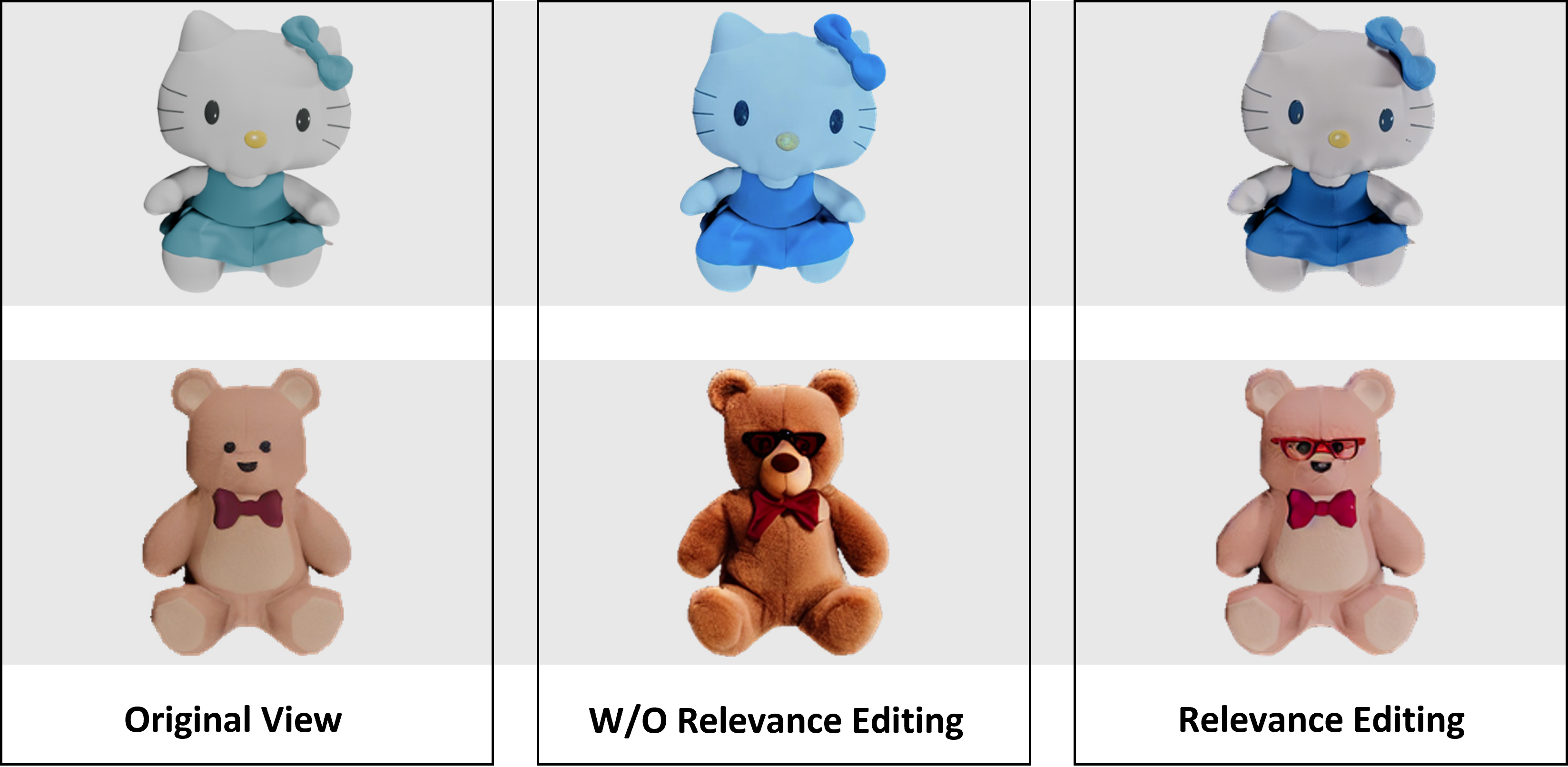}  
    \captionof{figure}{\textbf{Ablation Study of Relevance Editing.} Comparison of our image editing results with those of IP2P. For both models, $S_T$ and $S_I$ were set to 7.5 and 1.5, respectively. IP2P fails to isolate the specified region and tends to over-edit the input content. In contrast, our model accurately predicts the editing scope and confines the edits to the designated area.}
    \label{fig:relevence}
\end{figure}

\begin{figure*}
    \centering
    \includegraphics[width=\linewidth]{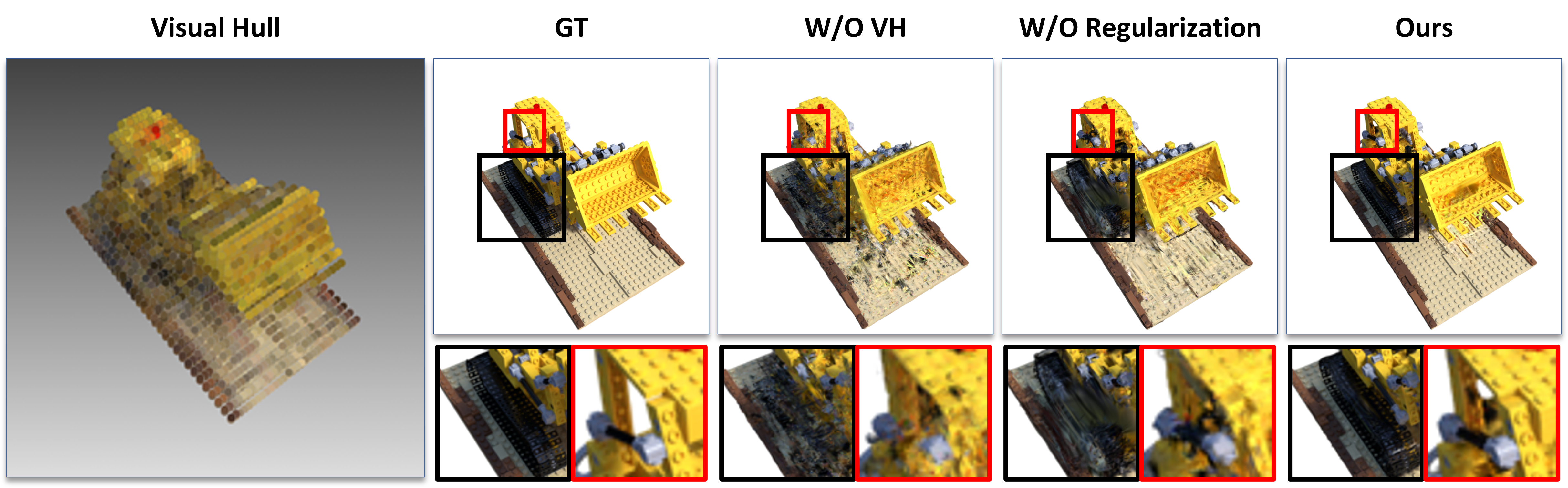}
    \caption{\textbf{Ablation Study on Various Components.} The term ``VH'' refers to the visual hull. It is evident that the inclusion of the visual hull enhances the completeness of the reconstructed object's structure. Moreover, the integration of depth regularization contributes to a richer and more refined representation of the object's details.}
    \label{fig:construction}
\end{figure*}

As illustrated in Fig. \ref{fig:relevence}, we conducted an ablation study to investigate the efficacy of the correlation editing strategy. In the absence of this strategy, methods such as IP2P \cite{brooks2023instructpix2pix} tend to modify the entire two-dimensional image, which results in suboptimal editing outcomes. However, the implementation of the correlation editing strategy, which restricts the pixels subject to editing, effectively mitigates this issue. Furthermore, we executed a series of experiments to evaluate the effectiveness of each component within the sparse reconstruction stage. These experiments were performed using the NeRF dataset with six input views, and the average metrics were subsequently reported. We systematically disabled the structural prior initialization and Gaussian depth regularization, one at a time, to assess their individual contributions. The results, presented in Table~\ref{tab:Ablation} and Fig. \ref{fig:construction}, indicate that each component significantly enhances performance, and their omission results in a deterioration of outcomes. 
\begin{table}[ht]
    \centering
    \setlength{\tabcolsep}{32pt}
    \caption{\textbf{Ablation study on key components.} The table shows the performance of our method (SVGS) with and without the visual hull and regularization components. Results highlight that both components contribute to improved LPIPS, PSNR, and SSIM scores, with the full method outperforming the ablated versions.}
    \begin{tabular}{lccc}
        \toprule
        Method & LPIPS\textsuperscript{*} $\downarrow$ & PSNR $\uparrow$ & SSIM $\uparrow$ \\
        \midrule
        Ours w/o Visual Hull & 0.18 & 20.08 & 0.7693 \\
        Ours w/o Regularization & 0.17 & 21.42 & 0.8134 \\
        \midrule
        \rowcolor{gray!30}SVGS (Ours) & \textbf{0.13} & \textbf{23.36} & \textbf{0.8497} \\
        \bottomrule
    \end{tabular}
    \label{tab:Ablation}
\end{table}
\subsection{Discussion}
Efficient and controllable 3D object editing represents a novel and significant area of research. Nevertheless, the technical specifics within this domain are still relatively underexplored. To address this gap, we developed an innovative framework that facilitates 3D object editing utilizing single-view input. Through a series of evaluative experiments, we demonstrated the efficacy of each module and successfully achieved text-driven editing outcomes that aligned with user expectations.
\\\indent The single-view editing strategy we propose effectively mitigates the distortion of three-dimensional reconstructions that can arise from inconsistencies in editing outcomes. By integrating structural prior initialization with depth regularization techniques, we have significantly enhanced the quality of models reconstructed from sparse views, leading to remarkable editing performance. This methodology not only improves editing accuracy but also guarantees the completeness and consistency of 3D reconstructions, thereby providing users with a more natural and precise experience in 3D object editing.
\begin{figure}
\centering
\includegraphics[width=\linewidth]{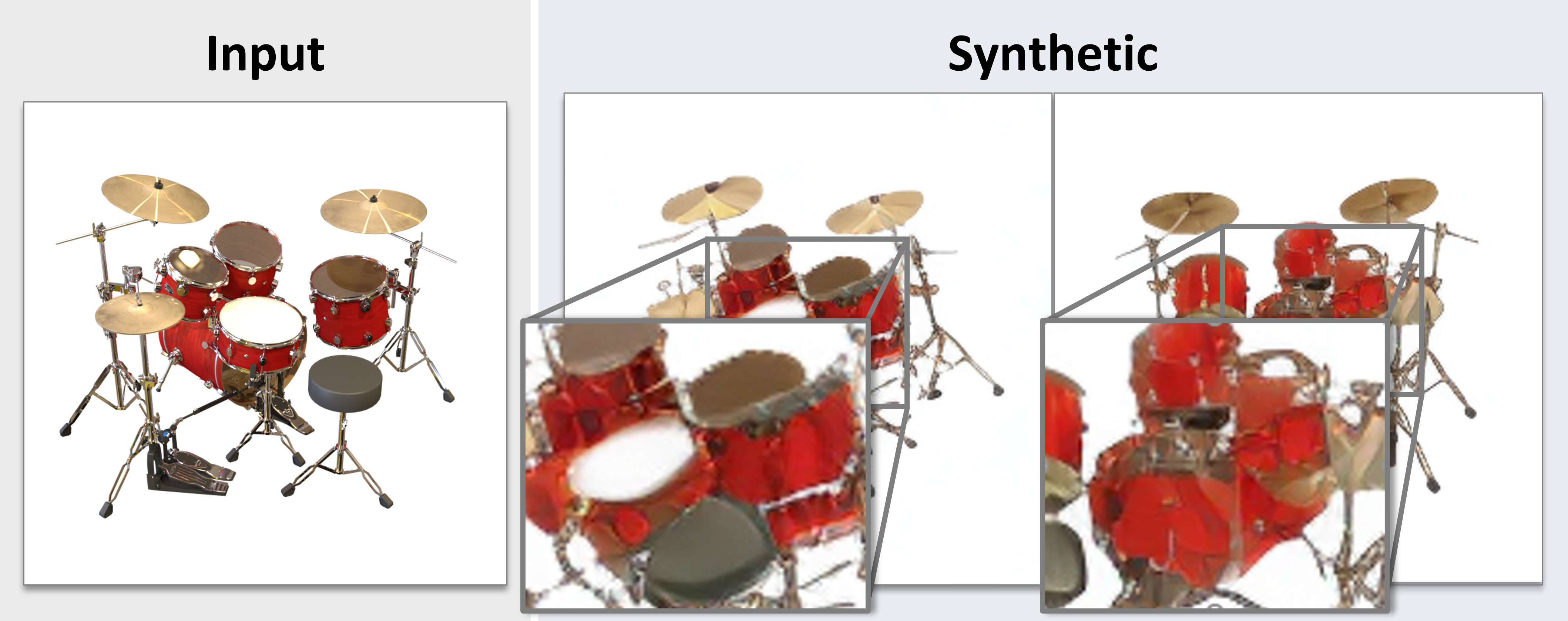}
\captionof{figure}{\textbf{Illustration of mutli-view diffusion model limitations.} 
For complex objects, particularly those composed of multiple components, the processing by UNet often results in multi-view images that struggle to maintain the original object's structure. This inconsistency in the generated structure can lead to a reconstructed model that fails to achieve the desired editing outcomes, leaving users dissatisfied.}
\label{fig:drawback}
\end{figure}
\\\indent However, our method is subject to certain limitations, primarily stemming from its dependence on the mutli-view diffusion model framework. As illustrated in Fig. \ref{fig:drawback}, a significant limitation arises when addressing complex objects; the mutli-view diffusion model encounters difficulties in generating multi-view images that align with the original 3D object structure. This inconsistency may result in suboptimal final rendering outcomes, particularly regarding the accuracy and coherence of multi-view image generation.
\\\indent Future research will concentrate on mitigating the identified limitations. We intend to integrate multi-view depth information to inform the mutli-view diffusion model, thereby facilitating the generation of consistent multi-view images and enhancing the precision and quality of the editing process. Furthermore, we will create new datasets to augment the capabilities of IP2P, thereby broadening its editing scope and addressing the existing constraints related to the diversity of scenes and objects it can accommodate. These advancements will establish a robust technical foundation for improving the overall effectiveness and applicability of 3D object editing.

	\section{Conclusion}
	In this study, we present SVGS, an innovative 3D editing algorithm that leverages mutli-view diffusion models and Gaussian stitching. This algorithm is specifically designed to enhance the controllability, consistency, and efficiency of 3D object editing. We introduce a relevance-based editing algorithm that facilitates fine-grained text editing through the use of IP2P. Furthermore, we propose a structural prior initialization method aimed at improving the quality of sparse view reconstruction. Empirical evaluations across multiple datasets indicate that SVGS exhibits significant superiority over alternative methods in terms of both editing quality and controllability.

\begin{acks}
This work was supported by the Natural Science Foundation of Zhejiang Province (No. LZ24F020001), the National Natural Science Foundation of China (No. 62576178), the Opening Foundation of the Tongxiang Institute of General Artificial Intelligence (No. TAGI2-B-2024-0009), and the State Key Laboratory of Advanced Medical Materials and Devices (No. SQ2022SKL01089-2025-14). This work was also supported by the program "Excellence initiative - research university" for the AGH University of Krakow, the ARTIQ project UMO-2021/01/2/ST6/00004 and ARTIQ/0004/2021, and by Polish Ministry of Science and Higher Education funds assigned to the AGH University of Krakow.
\end{acks}

\bibliographystyle{ACM-Reference-Format}
\bibliography{main}




\end{document}